\documentclass[runningheads]{llncs}

\PassOptionsToPackage{table,dvipsnames}{xcolor}
 
\usepackage[mobile]{eccv}

\usepackage{eccvabbrv}

\usepackage{graphicx}
\usepackage{booktabs}
\usepackage{wrapfig}
\usepackage{eso-pic}

\usepackage[accsupp]{axessibility}  

\usepackage{orcidlink}

\usepackage{booktabs}   
\usepackage{multirow}   
\usepackage{makecell}
\usepackage{booktabs}
\usepackage[table]{xcolor} 

\usepackage{amsmath,amsfonts,bm}

\def\eqref#1{equation~\ref{#1}}

\def\1{\bm{1}}

\DeclareMathAlphabet{\mathsfit}{\encodingdefault}{\sfdefault}{m}{sl}
\SetMathAlphabet{\mathsfit}{bold}{\encodingdefault}{\sfdefault}{bx}{n}

\definecolor{lasallegreen}{rgb}{0.03, 0.47, 0.19}
\definecolor{hanpurple}{rgb}{0.32, 0.09, 0.98}
\definecolor{green(pigment)}{rgb}{0.0, 0.65, 0.31}

\definecolor{GoodBlue}{HTML}{1F77B4}
\definecolor{BadRed}{HTML}{D62728}
\definecolor{BestGreen}{HTML}{2CA02C}

\newcommand{\red}[1]{{\color{red}#1}}

\newcommand{\gd}[1]{\textcolor{GoodBlue}{#1}}  
\newcommand{\bd}[1]{\textcolor{BadRed}{#1}}     

\newcommand{\citeps}[1]{{{\small\cite{#1}}}}

\newcommand{\cites}[1]{{{\small\cite{#1}}}}

\definecolor{Gray}{gray}{0.6}                        

\definecolor{cvprblue}{rgb}{0.21,0.49,0.74}
\hypersetup{
    colorlinks=true,
    linkcolor=red,
    citecolor=cvprblue,
    urlcolor=cvprblue,
    filecolor=cvprblue
}
\begin{document}

\title{Hybrid Event–Frame Sensors: Modeling, Calibration, and Simulation}

\titlerunning{Hybrid Event–Frame Sensors: Modeling, Calibration, and Simulation}

\author{Yunfan Lu$^{1,2}$, ~Nico Messikommer$^{2}$, ~Xiaogang Xu$^{3}$, ~Liming Chen$^{4}$, ~Yuhan Chen$^{1}$, ~Nikola Zubi\'{c}$^{2}$, ~Davide Scaramuzza$^{2~\star}$, ~Hui Xiong$^{1}$\thanks{Corresponding authors}
}

\authorrunning{Yunfan Lu et al.}

\institute{$^1$AI Thrust, HKUST(GZ) $^2$ RPG, University of Zurich $^3$ CUHK $^4$ AlpsenTek}

\AddToShipoutPictureFG*{\AtPageUpperLeft{\raisebox{-8mm}{\makebox[\paperwidth][c]{\footnotesize\color{gray}\shortstack[c]{This paper has been accepted for publication at the\\European Conference on Computer Vision (ECCV), 2026}}}}}
\maketitle

\begin{abstract}
Hybrid event–frame sensors integrate an Event Vision Sensor (EVS) and an Active Pixel Sensor (APS) within a single chip, combining the high dynamic range and low latency of the EVS with the rich spatial intensity information from the APS. 
While this tight integration offers compact, temporally precise imaging, the complex circuit architecture introduces non-trivial noise patterns that remain poorly understood and unmodeled.
In this work, we present the first unified, statistics-based imaging noise model that jointly describes the noise behavior of APS and EVS pixels.
Our formulation explicitly incorporates photon shot noise, dark current noise, fixed-pattern noise, and quantization noise, and links EVS noise to illumination level and dark current. 
Based on this formulation, we further develop a calibration pipeline to estimate noise parameters from real data and offer a detailed analysis of both APS and EVS noise behaviors.
Finally, we propose H-ESIM, a statistically grounded simulator that generates RAW frames and events under realistic, jointly calibrated noise statistics. 
Experiments on two hybrid sensors validate our model across multiple imaging tasks (e.g., video frame interpolation and deblurring), demonstrating strong transfer from simulation to real data. 
\url{https://yunfanlu.github.io/HESIM}
\end{abstract}
\vspace{-8pt}
\section{Introduction\label{sec:introduction}}
\vspace{-3pt}

\begin{figure}[t!]
\centering
\includegraphics[width=0.99\linewidth]{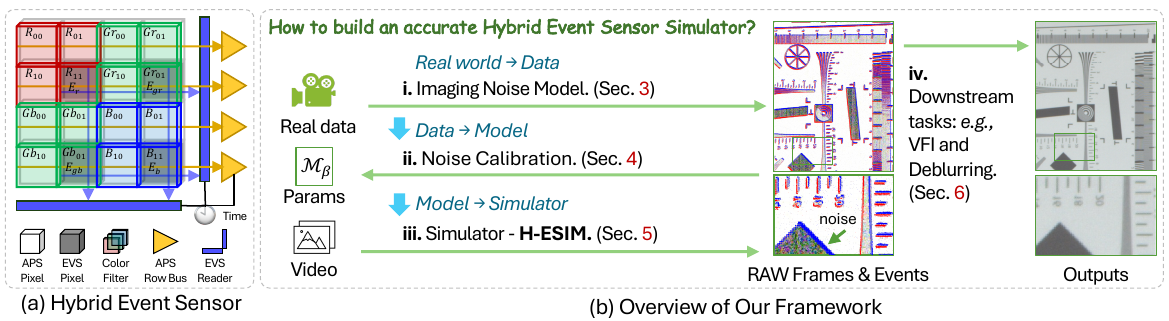}
\vspace{-6pt}
\caption{\small \textbf{(a)} Hybrid event-frame sensor with interleaved APS and EVS pixels, example Quad-Bayer layout. \textbf{(b)} Our framework: \textit{\textbf{i.}} a unified imaging noise model $\mathcal{M}_{\beta}$ for both APS and EVS (Sec.~\ref{sec:modeling}); \textbf{\textit{ii.}} calibration of APS/EVS noise (Sec.~\ref{sec:calibration}); \textbf{\textit{iii.}} Hybrid event sensor simulator, H-ESIM, which generates RAW frames and events (Sec.~\ref{sec:simulator}). \textbf{\textit{iv.}} downstream tasks, such as video frame interpolation, video deblurring. (Sec.~\ref{sec:downstream_tasks})\label{fig:1-CoverFigure-ICLR}}
\label{fig:1-CoverFigure}
\vspace{-12pt}
\end{figure}

Event cameras~\citeps{gallego2020event,wu2025mipi}, due to their asynchronous triggering mechanism, provide high dynamic range and low latency, enabling progress in vision tasks, \textit{e.g.,} high speed video imaging, and optical flow estimation~\citeps{gehrig2024low,yang2024vision}.
In contrast, frame-based cameras capture dense global intensity information, making the two modalities complementary in both time and space.
Consequently, many studies have explored event–image fusion to exploit these complementary advantages.
Nevertheless, most existing approaches achieve spatial alignment between images and events using beam splitters~\citeps{lin2023event,tulyakov2022time,sun2025low,zou2024eventhdr}, which require complex optical alignment and lead to bulky, impractical hardware designs.
Hybrid sensors \citeps{wu2025mipi,guo20233,kodama20231,yang2024vision,brandli2014240}
address these limitations by leveraging advanced semiconductor integration to output events and intensity frames from a \textbf{single} chip that integrates an Active Pixel Sensor (APS) and an Event Vision Sensor (EVS), as shown in Fig.~\ref{fig:1-CoverFigure-ICLR} (a).
Specifically, the APS captures dense intensity frames, while the EVS records pixel-wise brightness changes with microsecond latency. 
Their shared circuit architecture ensures precise spatiotemporal alignment.
This compact integration enables hybrid sensors to be easily applied to embedded platforms, \textit{e.g.,} mobile phones and drones \citeps{mipi_2024,MIPI2025hybridsensor,lu2024event}.
However, the increased circuit complexity of hybrid sensors introduces more complex noise patterns than those observed in conventional cameras \citeps{mipi_2024}.
\textit{This leads to an open problem: how to accurately calibrate hybrid sensor noise and develop a simulation framework that supports downstream vision tasks?}

To better characterize the noise distribution of hybrid sensors, we develop the \textbf{first} statistics-based imaging noise model that jointly describes the noise behavior of APS and EVS within a unified framework.
This model builds on prior works on standard cameras~\citeps{wei2020physics,johnson2024understanding}, while incorporating the circuit mechanisms of event pixels~\citeps{graca2021unraveling,graca2023optimal}.
In the hybrid sensor architecture, we explicitly model photon shot noise~\citeps{hasinoff2021photon}, dark current noise~\citeps{mcgrath2018dark}, fixed-pattern noise \citeps{georgiev2015fixed}, and quantization noise \citeps{schreier2005understanding}.
These noise sources exist in both APS and EVS. 
However, prior work rarely provides quantitative separation or measurement for EVS.
To address this, we first introduce the \textit{Q}-function \cites{isukapalli2008analytically,tanash2021improved} from statistics to link observed event probabilities with illumination and dark current.
As a result, the imaging noise model provides a unified explanation of how noise originates and propagates in both pixels, clarifying their correlations and differences.

Based on the unified noise model, we design a calibration pipeline to estimate key noise parameters from real data. 
APS and EVS are intrinsically linked through their shared scene luminance. 
Therefore, our approach derives direct and interpretable estimates through APS calibration, instead of relying on indirect brightness inference~\citeps{lin2022dvs}.
For APS, we employ multi-exposure compositing \citeps{song2023fixed} to recover cleaner intensity frames and adopt a statistical estimation method rather than a purely non-parametric fit~\citeps{mosleh2024non} to ensure interpretability. 
This step supports APS analysis and provides a clean brightness reference for EVS calibration.
Calibrating EVS is more challenging because its differential trigger outputs only discrete $\pm 1$ signals. 
In EVS, noise arises from photon shot and dark current drift, with rates that depend on both irradiance \citeps{cao2025noise2image} and bias settings \citeps{graca2021unraveling}. 
To overcome this, we employ an inverse $Q$-function formulation \citeps{tanash2020global,tanash2021improved,chiani2003new,karagiannidis2007improved} that relates observed brightness to noise event probabilities, enabling quantitative estimation of EVS noise parameters.
The calibrated parameters and noise models allow consistent APS–EVS noise generation and support the subsequent simulation.

Based on the calibrated parameters, we develop the hybrid sensor simulator, \textbf{H-ESIM}.
Given a high-frame-rate video, H-ESIM jointly generates RAW frames and events with calibrated noise statistics.
On the APS side, we design a controllable \emph{sRGB}$\rightarrow$\emph{RAW} inverse pipeline that injects calibrated fixed-patterns, shot, and dark–current noise, and we pair it with an ISP implemented in NumPy~\citeps{harris2020array}.
On the EVS side, in contrast to prior simulators~\citeps{rebecq2018esim,hu2021v2e}, which rely on hand–tuned hyperparameters and frame interpolation, H-ESIM is the \textbf{first} framework to unify APS and EVS simulation under a statistical model with parameters estimated from real data.
We explicitly apply the calibrated threshold and noise parameters to simulate the log–difference triggering process.
As a result, these components together yield a simulator that closely follows real hybrid sensors in both noise statistics and event triggering behavior.
Furthermore, to remove interpolation artifacts~\cites{hu2021v2e}, we introduce a high–speed video dataset ($3200$\,\textit{fps}) as input, which improves temporal fidelity.
Consequently, H-ESIM provides an open and reproducible implementation and produces consistent appearance between simulated and real data.

We evaluate the framework on two hybrid sensors with different pixel layouts and color filters.
The evaluation includes two stages.
First, we assess calibration accuracy by comparing estimated and measured noise distributions.
The results show close agreement with real sensor measurements.
Second, networks trained on our simulated data improve performance on real data.
This suggests that the calibrated simulation reduces the domain gap.
In summary, we present 
\textbf{(1)} the \textbf{first} comprehensive imaging noise model for hybrid event sensors, connecting shot, dark-current, fixed-pattern, quantization, and layout-dependent effects; 
\textbf{(2)} a quantitative calibration pipeline and dataset for estimating key noise parameters; 
\textbf{(3)} {H-ESIM}, the hybrid sensor simulator that generates RAW frames and events, enabling reproducible research and downstream vision applications.

\vspace{-3pt}
\section{Related Works}
\vspace{-3pt}
\noindent\textbf{Event Sensor:} 
Event cameras evolved from neuromorphic silicon retina concepts~\citeps{mead1988silicon} to the first dynamic vision sensor (DVS)~\citeps{lichtsteiner2008128_}.
The DVS encoded temporal contrast with microsecond latency and high dynamic range.
Later stream-based architectures~\citeps{posch2010qvga,berner2013240x180,brandli2014240} enhanced noise robustness, resolution, and efficiency, and stacked implementations \citeps{guo2017live,finateu20205} further reduced pixel pitch and maintained high HDR.
However, pure event sensors lack absolute intensity information.
To address this limitation, hybrid designs~\citeps{guo20233,kodama20231,yang2024vision} combine frames and events to exploit their complementary properties.
Hybrid architectures have recently attracted research interest, although they remain under active development~\citeps{mipi_2024,MIPI2025hybridsensor}.

\noindent\textbf{Standard Camera Noise:}
Noise is inherent in imaging systems. Accurate noise calibration is important for modeling image formation.
Recent work explores different strategies, including statistical models with parameter fitting~\citeps{wei2020physics} and data-driven approaches such as GAN or flow-based models that learn noise statistics from data~\citeps{chang2020learning,maleky2022noise2noiseflow,kousha2022modeling}.
Non-parametric methods~\citeps{zhang2021rethinking,mosleh2024non} reuse real noise patches without explicit assumptions.
Hybrid models combine physical priors with learning-based components to enhance robustness under extreme conditions~\citeps{cao2023physics,li2025noise}.
These methods provide useful insights. 
However, modeling hybrid APS noise remains challenging due to shared layout and circuit coupling.

\begin{figure*}[t]
\centering
\includegraphics[width=0.99\linewidth]{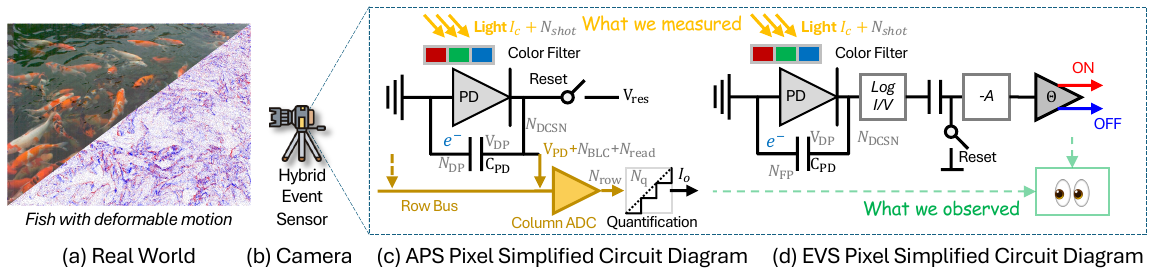}
\vspace{-6pt}
\caption{\small Hybrid sensor imaging pipeline and simplified pixel circuits. 
\textbf{(a)} Real scene.
\textbf{(b)} Hybrid event sensor. 
\textbf{(c)} APS path: photons filtered by the CFA generate charge in the photodiode, which is integrated and read through a shared row bus and column ADC.
The noise sources include photon shot noise $N_{\text{shot}}$, dark-current shot noise $N_{\text{DCSN}}$, black-level correction noise $N_{\text{BLC}}$, row noise $N_{\text{row}}$, and quantization noise $N_q$.
\textbf{(d)} EVS path: the photodiode drives a logarithmic stage, differentiator, and comparator with threshold $\Theta$ to emit events; dominant noise arises from photon shot and dark-current processes, fixed-pattern offsets.
The shared photoreceptor ensures tight temporal and spatial alignment, while the mixed layout introduces cross-coupled noise that our model and calibration explicitly address.}
\vspace{-16pt}
\label{fig:2-APS-Noise-Calibration-Pixel-Circuit}
\end{figure*}

\noindent\textbf{Event Camera Noise:}
Event noise differs from noise in frame-based sensors.
Early works investigated noise sources at the circuit level.
\citeps{nozaki2017temperature} quantified the influence of dark current, leakage, and parasitic photocurrent on background activity, while \citeps{graca2021unraveling} showed that the event photoreceptor behaves as a second-order system.
Furthermore, \citeps{graca2023optimal} identified a physical noise floor, showing that event noise is lower-bounded by twice the photon shot noise. 
At the statistical level, ESIM~\citeps{rebecq2018esim} and v2e~\citeps{hu2021v2e} introduced random contrast thresholds to emulate noisy triggering.
\citeps{lin2022dvs} proposed a Brownian-motion-based noise model.
More recently, \citeps{cao2025noise2image} related event noise to photon arrival statistics, revealing that noise events encode pixel intensity information.
\citeps{oliver2025event} characterized background activity under controlled illumination, emphasized the need for physics-grounded noise models.
Additionally, \citeps{shiba2025simultaneous} proposed an event denoising method, yet still acknowledged that explicit noise modeling remains challenging.

\noindent\textbf{Event Camera Simulators:}
Simulators play an important role in event vision. They enable training and evaluation in scenarios that are difficult to capture.
Rule-based methods, \textit{e.g.,} ESIM~\citeps{rebecq2018esim}, v2e~\citeps{hu2021v2e}, DVS-Voltmeter~\citeps{lin2022dvs}, ADV2E~\citeps{jiang2024adv2e}, and PECS~\citeps{han2024physical}, generate events using handcrafted rules to 3D scenes or high-speed videos.
Recent variants~\citeps{zhang2024v2ce,tsuji2023event,amini2022vista} improve physical fidelity by integrating temporal continuity, spectral rendering, or multi-modal integration.
In contrast, learning-based methods~\citeps{zhu2021eventgan,pantho2022event} learn sensor noise characteristics from data but offer limited interpretability.
Despite these advances, existing simulators do not provide a unified framework for joint APS–EVS modeling. In particular, they do not couple the two modalities through shared and calibrated parameters.

\vspace{-3pt}
\section{Statistical Imaging Noise Model\label{sec:modeling}}
\vspace{-9pt}
APS and EVS of hybrid sensors share the same optical system, \textit{e.g.,} the lens and aperture.
Therefore, the ideal electrical signal $I_c(t;\Delta t)$ at time $t$ is shared across both modalities, as shown in Fig.~\ref{fig:2-APS-Noise-Calibration-Pixel-Circuit}.
This signal is obtained by integrating the scene radiance $L(\tau, \lambda)$ filtered by the color filter $C_f(\lambda)$ and photodiode circuit $P_c(\lambda)$ over the photodiode integration time $[t\!-\!\Delta t,t]$ and visible wavelength range $\Lambda$ defined in Eq.\ref{eq:light_to_intensity}.
\begin{equation} \small
I_{c}(t;\Delta t) = \int_{t-\Delta t}^{t} \int_{\Lambda} P_c(\lambda) C_f(\lambda)  L(\tau, \lambda) d\lambda\ d\tau
\label{eq:light_to_intensity}
\end{equation}
The APS and EVS pixels differ in how they sample this shared signal $I_c$.
The APS pixel performs integral sampling over $\Delta t$, producing RAW intensity frames, whereas the EVS pixel applies differential, thresholded readout of the \textit{logarithmic} signal to emit events.
Therefore, both modalities can be modeled within a single physics-informed statistical framework that takes $I_c(t;\Delta t)$ as the shared input while accounting for the APS exposure time and the EVS sensitivity.
This unified view enables the separation of illumination-dependent, exposure time-dependent, and fixed noise terms, and reveals cross-coupled effects from the mixed layout and shared readout chain.
In the following, we detail the APS and EVS noise models and their interconnections.

\noindent\textbf{APS Pixel:}
Since APS directly records the integrated intensity, we model its output $I_o$ as the ideal signal $I_c$ plus three additive noise components, as expressed in Eq.~\ref{eq:aps_noise_groups}: illumination-dependent shot noise $N_\text{shot}$, exposure time-dependent noise $N_\text{time}$, and fixed noise $N_\text{fixed}$.
These noise sources are treated as approximately additive terms, and following previous studies~\citeps{chang2020learning,cao2023physics,luisier2010image}, each is modeled using Gaussian or uniform distributions for tractability.
First, photon shot noise arises from Poisson photon arrivals~\citeps{joiner1971some,wei2020physics}.
Therefore, in statistical signal modeling, when the photon count per pixel is moderate or high, the Poisson distribution can be approximated by a Gaussian distribution~\citeps{foi2008practical,boncelet2009image}.
Second, temporal noise combines a linear drift from defective, hot or leaky, pixels with dark-current shot noise, whose variance increases with exposure time and temperature \citeps{johnson2024understanding}.
Third, fixed noise $N_{\text{FP}}$ includes spatially structured offsets, row noise $N_{\text{row}}$, and black-level correction $N_{\text{BLC}}$.
Hybrid layouts amplify $N_{\text{row}}$, making row noise more pronounced.
Additional contributions include read noise and quantization noise, which we model as uniform with step size $q$.
Finally, a residual term $N_\text{res}$ aggregates small unmodeled effects and is used only for completeness.
\begin{equation} \footnotesize
\begin{aligned}
I_o &\!=\!I_c 
  \!+\! \underbrace{N_\text{shot}(I_c)}_{\substack{\text{illum.}~\text{dep.}}}
  \!+\! \underbrace{N_\text{time}(\Delta t)}_{\substack{\text{expo.}~\text{dep.}}}
  \!+\! \underbrace{N_\text{fixed}}_{\text{constant}} \\
&\!=\! I_c 
  \!+\! \underbrace{\mathcal{N}\bigl(0,\sigma^2_{\text{shot}}\bigr)}_{N_\text{shot}(I_c)}
  \!+\! \underbrace{\Delta t\, N_{\text{DP}} 
    \!+\! \mathcal{N}\bigl(0,\sigma^2_{\text{DCSN}}\bigr)}_{N_\text{time}(\Delta t)} 
    \!+\! \underbrace{N_{\text{FP}} +\mathcal{N}\bigl(0,\sigma^2_{\text{read}}\bigr)
  \!+\! \mathcal{U}\bigl(\frac{-1}{2q},\!\frac{1}{2q}\bigr)}_{N_\text{fixed}} 
  \!+\! N_\text{res}
\label{eq:aps_noise_groups}
\end{aligned}
\end{equation}
\noindent\textbf{EVS Pixel:}
The event pixel first accumulates charge, which in turn modulates the voltage, as illustrated in Fig.~\ref{fig:2-APS-Noise-Calibration-Pixel-Circuit} (d).
Between two consecutive times $t_0$ and $t_1$, an event is triggered only when the voltage difference at \textit{logarithmic} domain between $V_{t_0}$ and $V_{t_1}$ exceeds the predefined threshold $\theta$.
This mechanism makes the event generation process more complex than that of the APS.
In hybrid sensors, the EVS operates at a fixed sampling rate to eliminate the \textit{scanning pattern} observed in earlier designs, \textit{e.g.}, DVS346~\cites{DAVIS346AER_2023,yates2020robotic}.
At each recording time, the pixel integrates photocurrent over a short interval.
The ideal accumulated voltage is denoted as $\hat V_t$. It is linearly proportional to the integrated photocurrent $I_c$, following the charge–voltage relation in the pixel circuit.
The observed voltage is corrupted by multiple noise sources, photon shot noise $N_{\text{shot}}$, dark-current shot noise $N_{\text{DCSN}}$, fixed offsets $N_{\text{FP}}$, and a baseline voltage $V_{\text{PD}}$ from the photodiode circuit as in Eq.~\ref{eq:event_with_noise}.
\begin{equation} 
V_{t_1} = \hat V_{t_1}
            + N_{\text{shot}}
            + N_{\text{DCSN}} 
            + N_{\text{FP}}
            + V_{\text{PD}}
\label{eq:event_with_noise}
\end{equation}
For two consecutive samples $t_0$ and $t_1$, the voltage difference is defined by Eq.~\ref{eq:delta_v_k_with_taly}.
At time $t_1$, an event is generated as $\mathcal{Q}_e(\Delta V_{t_1})$, where $\mathcal{Q}_e(\cdot)$ denotes the quantization function: an ON event ($+1$) is triggered when $\Delta V_{t_1} > \theta$, an OFF event ($-1$) is triggered when $\Delta V_{t_1} < -\theta$.
To analyze the noise distribution, we decompose $\Delta V_{t_1}$ into two parts as shown in Eq.~\ref{eq:delta_v_k_with_taly}: the \emph{signal term} $S$, determined by the clean voltages $\hat V_{t_0}, \hat V_{t_1}$, and the \emph{noise term} $N_e$, produced by various noise sources.
Here, the instantaneous noise at time $t$ is given by $n_t = N_{\text{shot}} + N_{\text{FP}} + N_{\text{DCSN}}$.
\begin{equation} \small
\begin{aligned}
\Delta V_{t_1} 
&= \log V_{t_1} - \log V_{t_0} \\
&= \underbrace{\log({1 + \frac{n_{t_1}}{\hat V_{t_1} + V_{\text{PD}}}}) - \log({1 + \frac{n_{t_0}}{\hat V_{t_0}+V_{\text{PD}}}})}_{\text{Noise~part, $N_e$}} 
+ \underbrace{\log(\frac{\hat V_{t_1} + V_{\text{PD}}}{\hat V_{t_0}+V_{\text{PD}}})}_{\text{Signal~part, $S$}}
\end{aligned}
\label{eq:delta_v_k_with_taly}
\end{equation}
Based on this formulation, we derive the relationship between individual noise components and the voltage difference.
Under non-extreme low-light conditions, where noise magnitude is smaller than the signal level, this relationship can be approximated using a first-order Taylor expansion.
Accordingly, we define the event noise $N_e$ as in Eq.~\ref{eq:delta_t_1}.
The noise components are modeled as follows.
Photon shot noise $N_{\text{shot}}$ depends on irradiance.
It is approximated as $N_{\text{shot}} \sim \mathcal{N}(0,\sigma^2_{\text{shot}})$, where $\sigma^2_{\text{shot}} \propto \hat V_t$.
Dark-current shot noise is similarly approximated as $N_{\text{DCSN}} \sim \mathcal{N}(0,\sigma^2_{\text{DCSN}})$.
Fixed-pattern noise $N_{\text{FP}}$ arises from device mismatches and is treated as a deterministic offset.
\begin{equation} \small
N_e \approx 
        \frac{N_{\text{shot},t_1} + N_{\text{DCSN},t_1} + N_{\text{FP}}}{{\hat V_{t_1} + V_{\text{PD}}}}
    -
        \frac{N_{\text{shot},t_0} + N_{\text{DCSN},t_0} + N_{\text{FP}}}{{\hat V_{t_0} + V_{\text{PD}}}}
\label{eq:delta_t_1}
\end{equation}
Following previous work based on the Brownian-motion assumption~\citeps{lin2022dvs}, the event noise $N_e$ at $t_0$ and $t_1$ is assumed to be independent.
Accordingly, $N_e$ is modeled as a Gaussian distribution in Eq.~\ref{eq:event-sigma}, where we denote $\hat V_{t_1}+V_{\text{PD}}$ as $\mathbf{V}_{t_1}$ and $\hat V_{t_0}+V_{\text{PD}}$ as $\mathbf{V}_{t_0}$.
This Gaussian distribution of event noise $N_e$ has variance $\sigma^2_n$ and mean $\mu_n$.
For defective pixels, $\mu_n$ takes non-zero values.
This formulation explicitly shows how illumination conditions and circuit parameters jointly determine the probability of event noise.
\begin{equation} \small
\begin{aligned} 
\mathcal{N}\big(
\underbrace{{N_{\text{FP}}}\big(\frac{1}{\mathbf{V}_{t_1}}\!-\!\frac{1}{\mathbf{V}_{t_0}}\big)}_{\mu_n},\underbrace{{(\sigma_{\text{shot}}^2\!+\!\sigma_{\text{DCSN}}^2\!-\!2\rho\sigma_{\text{shot}}\sigma_{\text{DCSN}})}(\frac{1}{\mathbf{V}_{t_1}^2}\!+\!\frac{1}{\mathbf{V}_{t_0}^2}
}_{\sigma^2_n}
)\big)
\end{aligned}
\label{eq:event-sigma}
\end{equation}
Subsequently, we employ the $Q$-function \citeps{chiani2003new,tanash2020global,tanash2021improved} as defined in Eq.~\ref{eq:q_function}, to analyze the probability of event triggering.
Specifically, we denote the probabilities of ON / OFF events as $P_{+}$ and $P_{-}$, which can be expressed as Eq.~\ref{eq:positive_negative_possibilities}.
In the static scene case $S=0$, and zero-mean noise $\mu_n=0$, these probabilities simplify to $P_{+}=Q(\theta/\sigma_n)$ and $P_{-}=1-Q(-\theta/\sigma_n)=Q(\theta/\sigma_n)$.
\begin{equation} \small
Q(x) =\frac{1}{\sqrt{2\pi}}\int_{x}^{+\infty}e^{-t^2/2}dt 
\label{eq:q_function}
\end{equation}
\begin{equation} \small
\begin{aligned}
P_{+} & = \operatorname{P}(\Delta V_{t_1} > \theta) 
& = & \;\;Q\big(\frac{\theta - (S+\mu_n)}{\sigma_n}\big), \\
P_{-} & = \operatorname{P}(\Delta V_{t_1} < -\theta) 
& = & \;\;1 - Q\big(\frac{- \theta - (S+\mu_n)}{\sigma_n}\big)
\end{aligned}
\label{eq:positive_negative_possibilities}
\end{equation}
In summary, this statistical model establishes explicit links between the observed signals, RAW frame $I_o$ for APS frames and probabilities $P_{+}, P_{-}$ for EVS events, and their underlying noise sources, as formulated in Eq.~\ref{eq:aps_noise_groups} and Eq.~\ref{eq:positive_negative_possibilities}.
This unified viewpoint forms the foundation for noise calibration.

\vspace{-3pt}
\section{Hybrid Event Sensor Noise Calibration\label{sec:calibration}}
\vspace{-3pt}
To calibrate noise in a hybrid sensor, we first collect and pre-process a dedicated dataset, then model the noise of APS and EVS sequentially.
We adopt a control-variable strategy to estimate each group.
For APS pixels, we vary the exposure time under a fixed scene, 
whereas for event pixels, we control pixel brightness through scene design.
The calibration dataset includes two configurations:
(\textbf{i}) dark measurements with the lens fully covered, and
(\textbf{ii}) illuminated measurements with static multi-brightness patterns, \textit{e.g.,} checkerboards, color checker, resolution charts, each captured under multiple exposure times.

\noindent\textbf{APS Noise Calibration:}
Our goal is to estimate the parameters in Eq.~\ref{eq:aps_noise_groups}. We separate the composite noise $N_{a}$, and express the aggregate form as Eq.~\ref{eq:aps_noise_items}, where $\sigma^2_{\times}$ is the cross-term, which corrects variance among different noise sources.
\begin{equation} \small
\begin{aligned}
N_{a}\!=\!\mathcal{N}(0,\sigma^2_{\text{shot}}\!+\!\sigma^2_{\text{DCSN}}\!+\!\sigma^2_{\text{read}}\!-\!\sigma^2_{\times})\!+\!\Delta t N_{\text{DP}}\!+\!N_{\text{FP}} + N_{\text{q}}
\end{aligned}
\label{eq:aps_noise_items}
\end{equation}
The dark frames are $\mathbf{I_d}\!=\!\{I_d^{\Delta t_i}\}_{i=1}^T$
and the illuminated frames are
$\mathbf{I_w}\!=\!\{I_w^{\Delta t_i}\}_{i=1}^T$, where $T$ denotes the set of exposure times.
For each exposure time, we record $N_{\text{cal}}$ frames and denote the per-exposure means as $\bar{I}_{d}^{\Delta t}$ and $\bar{I}_{w}^{\Delta t}$.
Using the dark means and their linear dependence on exposure, we solve the relation
${\bar{I}_{d}^{\Delta t} = \Delta t N_{\text{DP}} + N_{\text{FP}}}$
to obtain $N_{\text{DP}}$ and $N_{\text{FP}}$.
We further decompose the fixed-pattern component as 
$N_{\text{FP}}\!=\!N_{\text{row}} + N_{\text{BLC}}$, where 
$N_{\text{row}}\!\in\!\mathbb{R}^{H}$ is the row bias and 
$N_{\text{BLC}}\!\in\!\mathbb{R}^{H\times W}$ is the per-pixel black-level component.
For illuminated frames, we subtract the calibrated dark components from $\bar{I}_{w}^{\Delta t}$ to obtain a bias-corrected estimate $\hat{I}_w^{\Delta t}\!=\!\bar{I}_{w}^{\Delta t}-\bar{I}_{d}^{\Delta t}$.
We then compute the empirical mean and variance of $\hat{I}_w^{\Delta t \text{~ms}}$ to estimate the sum of stochastic variances $\sigma^2_{\text{shot}}\!+\!\sigma^2_{\text{DCSN}}\!+\!\sigma^2_{\text{read}}$.
Specifically, taking the variance on both sides of Eq.~\ref{eq:aps_noise_items} yields
$\text{Var}(N_{a})\!=\!\sigma^2_{\text{shot}}\!+\!\sigma^2_{\text{DCSN}}\!+\!\sigma^2_{\text{read}}\!-\!\sigma^2_{\times}\!+\! 1/(12q^2))$.
We treat $I_c$ and $\Delta t$ as regressors and fit the variance by least squares.
We adopt the second-order polynomial model in Eq.~\ref{eq:order_2_var} and fit each pixel position within a Quad-Bayer block (16 positions) to obtain $\bm{\beta}_a\!=\!\{\beta_i\}_{i=0}^5$.
The second-order polynomial provides a flexible empirical model for residual signal- and exposure-dependent variance.
Photo response non uniformity (PRNU) is not separately identified in the current calibration; its effect could be absorbed by the per-position coefficients.
\begin{equation} \small
\begin{aligned}
\operatorname{Var}(N_{a})\!=\! \sigma^2_{\text{shot}}\!+\!\sigma^2_{\text{DCSN}}\!+\!\sigma^2_{\text{read}}\!-\!\sigma^2_{\times} 
\!=\!\beta_0\!+\!\beta_{1} I_c\!+\!\beta_{2}\Delta t\!+\!\beta_{3}I_c^2\!+\!\beta_4 I_c \Delta t\!+\!\beta_5 \Delta t ^ 2
\end{aligned}
\label{eq:order_2_var}
\end{equation}
\textit{\textbf{APS Noise Parameters:}}
The final calibration includes sixteen per-position coefficients $\bm{\beta}_a$ and fixed components $N_{\text{BLC}},N_{\text{row}},N_{\text{DP}}$.
This per-position calibration effectively captures spatial non-uniformity introduced by the hybrid layout while preserving a simple and interpretable mapping from $(I_c, \Delta t)$ to noise variance, as defined in Eq.~\ref{eq:aps_noise_items}, Eq.~\ref{eq:order_2_var}.

\noindent\textbf{Event Noise Calibration:}
The EVS noise calibration is to link observed noise events to the model parameters.
Under the threshold‐comparator model, the event probabilities are given by the $Q$ function, see Eq.~\ref{eq:q_function}, Eq.~\ref{eq:positive_negative_possibilities}.
In a static scene ($S=0$) and with zero‐mean bias ($\mu_n=0$), we obtain $P_+=P_-=Q(\theta/\sigma_n)$, which yields Eq.~\ref{eq:event_sigma_final_noise}.
This equation relates the unobserved noise variance $\sigma^2_n$ to the observable event probability $P$ and to $\hat V_t$, which is mapped from $I_c$.
\begin{equation} \small
\sigma^2_{n}
=\frac{2(\sigma_{\text{shot}}^2 + \sigma_{\text{DCSN}}^2 - 2\rho\sigma_{\text{shot}}\sigma_{\text{DCSN}})}{(\hat V_{t_1}+V_{\text{PD}})^2}
=(\frac{\theta}{Q^{-1}(P)})^2 \label{eq:event_sigma_final_noise}
\end{equation}
However, two main challenges arise:
\textbf{(1)} \textit{Domain mapping:} APS provides digital intensity $I_c$, whereas EVS operates in a logarithmic voltage domain ($\hat V_t$).
An interpretable, monotonic mapping $I_c\rightarrow\hat V_t$ is therefore required to bridge the two domains.
\textbf{(2)} \textit{Noise coupling:} Photon shot noise and dark‐current noise are correlated on a hybrid layout, which directly influences the estimation of $\sigma^2_n$.
To address these challenges, we collect noise events under consistent conditions, including dark scenes and static grayscale ramp scenes.
The APS calibration is applied to recover a clean pixel intensity $I_c$ by Eq.~\ref{eq:aps_noise_groups}.
Subsequently, an affine mapping projects APS intensity to the EVS voltage domain, $\hat V_t+V_{\text{PD}}\approx\beta_1 I_c+\beta_2$, under monotonicity and positivity constraints.
A scale factor $\beta_0$ converts the threshold units ensuring that $\beta_0\Theta$ is comparable to $Q^{-1}(P)$, here $\Theta$ is the hardware threshold, \textit{e.g.,} 0.75 \textit{mV}.
For random noise, we adopt an intensity‐dependent model consistent with the APS: $\sigma_{\text{shot}}=\beta_3 I_c$, $\sigma_{\text{DCSN}}=\beta_4$, and $\rho=-\beta_5$, which yields the regression form in Eq.~\ref{eq:event_estimation_final_noise}.
For the events with $N$ trials, we count $n_+,n_-$ and obtain
$P_+=n_+/N, P_-=n_-/N, P=\tfrac12(P_+ + P_-)$, then define the observation $Q^{-1}(P)$.
This procedure estimates parameter set $\bm{\beta}_e=\{\beta_i\}_{i=0}^5$ and produces a bad‐pixel mask, handling $\mu_n\neq0$ or outliers by separate modeling or removal.
\textit{\textbf{EVS Noise Parameters:}}
We first estimate $\mu_n$ from events captured under dark conditions, where signal-dependent components are minimized. This term represents the fixed offset.
Since Eq.~\ref{eq:event_estimation_final_noise} is nonlinear, we optimize the parameters using gradient descent to fit the relationship between pixel intensity $I_c$ and the observed event probability $P$.
After estimating $\bm{\beta}_e$, we compute the probability of events for each pixel under the current intensity using Eq.~\ref{eq:positive_negative_possibilities}.
Finally, we sample ON and OFF events according to these probabilities to produce the simulated event stream.
\begin{equation} \small
\begin{aligned}
Q^{-1}(P) = \frac{\theta}{\sigma_n}
&= \frac{\theta(\hat V_{t_1}+V_{\text{PD}})}{\sqrt{2(\sigma_{\text{shot}}^2 + \sigma_{\text{DCSN}}^2 - 2\rho~\sigma_{\text{shot}}~\sigma_{\text{DCSN}})}} \\
&= \frac{\beta_0  \Theta (\beta_1 I_c + \beta_2)}{\sqrt{2((\beta_3 I_c)^2 + \beta_4^2 + 2\beta_5  \beta_3 I_c  \beta_4)}}
\end{aligned}
\label{eq:event_estimation_final_noise}
\end{equation}

\begin{figure*}[t!]
\centering
\includegraphics[width=0.99\linewidth]{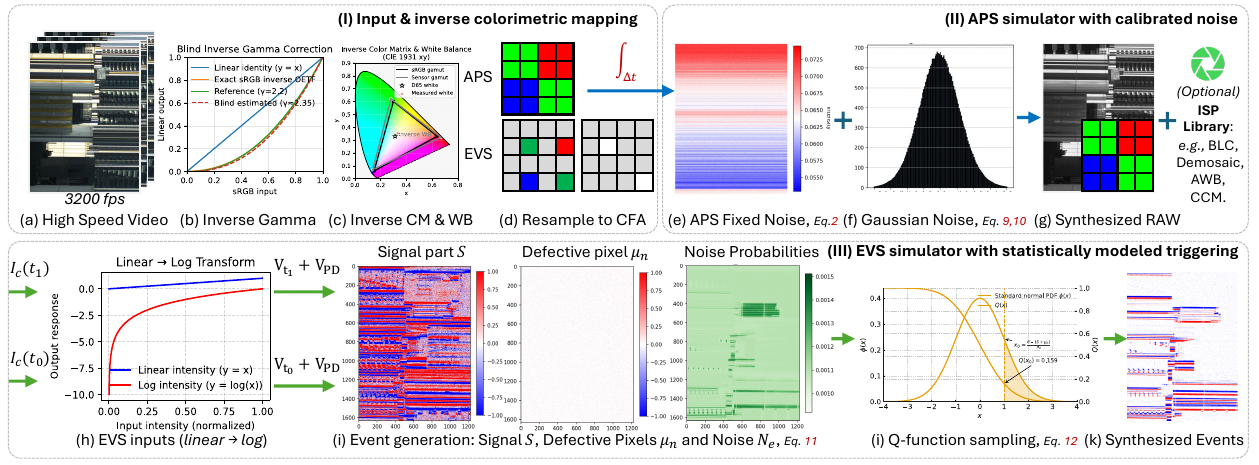}
\vspace{-7pt}
\caption{\small \textit{Key steps of \emph{H-ESIM}.}
(I) \textit{Input and inverse colorimetric mapping:} (a)–(d) map a 3200 \textit{fps} video to per-pixel intensity $I_c$ and APS/EVS CFA through inverse gamma and inverse color matrix and white balance.
(II) \textit{APS simulator with calibrated noise:} (e)–(g) add fixed terms $N_{\text{row}}$, $N_{\text{BLC}}$, and $\Delta t N_{\text{DP}}$, then sample the illumination/exposure–dependent variance $f_a(I_c,\Delta t;\bm{\beta}—a)$ to synthesize RAW (Eq.~\ref{eq:aps_noise_groups}, Eq.~\ref{eq:aps_noise_items}, Eq.~\ref{eq:order_2_var}). 
(III) \textit{EVS simulator with statistically modeled triggering:} (h)–(k) map $I_c$ to voltage, form the signal $S$ and noise $N_e$ with offsets $\mu_n$, compute $P_+$ and $P_-$ through $Q(\cdot)$ and threshold $\theta$ (Eq.~\ref{eq:event_sigma_final_noise}, Eq.~\ref{eq:event_estimation_final_noise}), and sample events using $\bm{\beta}_e$.}
\vspace{-10pt}
\label{fig:placeholder}
\end{figure*}

\vspace{-3pt}
\section{Hybrid Event Sensor Simulator (H-ESIM)\label{sec:simulator}}
\vspace{-3pt}
This section presents the design of our proposed hybrid event sensor simulator \emph{H-ESIM} based on the preceding statistical noise model and calibration.
Given a high–frame–rate video sequence, H-ESIM generates RAW frames and events whose noise follows the calibrated distributions.
Our objectives are: 
\textbf{(i)} to ensure interpretable and controllable noise that matches hardware behavior, and
\textbf{(ii)} to achieve efficient execution for both training and evaluation.
To eliminate interpolation artifacts, we leverage a high–speed dataset ($3200$ \textit{fps},$1296\times 1024$), which prevents the temporal distortions observed in previous pipelines~\citeps{rebecq2018esim,hu2021v2e}.

\noindent\textbf{APS Simulator:}
The synthesized RAW images should match the noise statistics of real RAW and, after the ISP, reproduce the appearance of the input \emph{sRGB} frames.
To achieve this, we implement a controllable \emph{sRGB}$\rightarrow$\emph{RAW} inverse pipeline and inject calibrated noise through the following steps:
\textbf{(1)} \textit{Inverse gamma correction:} convert RGB frames to a linear irradiance domain.
\textbf{(2)} \textit{Inverse color matrix and white balance:} map linear RGB to the camera color space and undo white balance.
\textbf{(3)} \textit{Resampling to Bayer:} rearrange into a single–channel RAW pattern according to a CFA, \textit{e.g.,} Quad–Bayer.
\textbf{(4)} \textit{Fixed terms:} following Eq.~\ref{eq:aps_noise_groups}, add $N_{\text{FP}}\!=\!N_{\text{row}}\!+\!N_{\text{BLC}}$ and $\Delta t N_{\text{DP}}$.
\textbf{(5)} \textit{Illumination/exposure–dependent noise:} using Eq. \ref{eq:order_2_var} and the position–dependent coefficients $\bm{\beta_a}$, sample the random terms driven by Eq.~\ref{eq:aps_noise_items}, Eq.~\ref{eq:order_2_var}.
\textbf{(6)} \textit{Quantization:} apply the camera bit depth to generate the synthesized RAW frames.

\begin{figure*}[t!]
\centering
\includegraphics[width=0.97\linewidth]{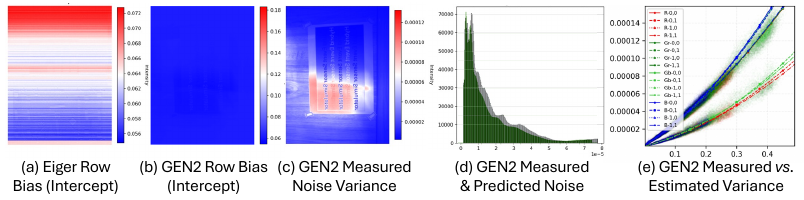}
\vspace{-8pt}
\caption{\small 
APS noise calibration on Eiger and GEN2.
(a–b) Estimated row bias (intercept).
(c) Measured per-pixel noise variance (GEN2).
(d) Distribution of measured and predicted noise.
(e) Measured vs. estimated variance across pixels. Scatter points denote measured noise variance, and curves denote fitted models.
}
\vspace{-10pt}
\label{fig:3-APS-Noise-Calibration-Light}
\end{figure*}

\noindent\textit{\textbf{APS ISP Library:}}
To provide a reproducible imaging pipeline, we implement an open ISP that connects the \emph{sRGB} and \emph{RAW} domains.
The implementation is provided in NumPy and includes black-level correction~\citeps{besuijen2006black}, demosaicing~\citeps{li2008image}, color correction~\citeps{xu2015vector}, gamma and tone mapping~\citeps{lee2011local}.
The ISP performs \emph{RAW}$\rightarrow$\emph{sRGB} conversion to maintain consistency with the inverse pipeline.
Compared to prior work~\citeps{yunfan2024rgb} implemented in MATLAB, our open implementation allows inspection and modification of the processing steps.
\textit{See Appx.~\red{C} for more details.}

\noindent\textbf{EVS Simulator:}
On the EVS pipeline, we explicitly apply the calibrated threshold, dark current and photon shot params $\bm{\beta_e}$, as well as position dependence and readout coupling , to drive the log-domain differencing and comparator process, as expressed in Eq.~\ref{eq:delta_v_k_with_taly} and Eq.~\ref{eq:positive_negative_possibilities}.
The workflow consists of the following steps:
\textbf{(1)} \textit{Linear intensity to voltage:} reuse APS steps 1–3 to obtain per–pixel $I_c$; map to the EVS voltage domain through the affine model $\hat V_t+V_{\text{PD}}\approx\beta_1 I_c+\beta_2$.
\textbf{(2)} \textit{Log differencing and sampling:} at times $t_0,t_1$ compute the signal term $S$ and the noise term $N_e$ of $\Delta V_{t_1}$ as defined in Eq.~\ref{eq:delta_v_k_with_taly}.
\textbf{(3)} \textit{Threshold and noise injection:} convert the hardware threshold $\Theta$ to the simulation threshold $\theta$; sample $N_e\sim\mathcal{N}(\mu_n,\sigma_n^2)$ using Eq.~\ref{eq:event_sigma_final_noise} and Eq~\ref{eq:event_estimation_final_noise} with $\sigma_{\text{shot}} \approx \beta_3 I_c$, $\sigma_{\text{DCSN}} \approx \beta_4$, and correlation $\rho \approx -\beta_5$.
\textbf{(4)} \textit{Trigger decision:} 
form $\Delta V_{t_1}=S+N_e$; emit ON if $\Delta V_{t_1}>\theta$, OFF if $\Delta V_{t_1}<-\theta$, else suppress; ensure the empirical trigger rates match Eq.~\ref{eq:positive_negative_possibilities}.
Here, $N_e$ includes photon shot noise, dark current, and position-dependent offsets, as defined in Eq.~\ref{eq:delta_v_k_with_taly} and Eq.~\ref{eq:event-sigma}.

\noindent\textbf{Implementation:}
We implement H-ESIM in NumPy \citeps{harris2020array} and PyTorch \citeps{paszke2019pytorch}.
The APS and EVS simulators share the same input and parameter interface, ensuring portability and cross-verification.
The simulator also supports parameter modification for quantitative noise analysis and the simulation of more complex noise patterns.
It reproduces the log-difference triggering mechanism and realistic noise behavior, providing stable and statistically consistent data for training and evaluating downstream tasks.

\begin{figure*}[t!]
\centering
\vspace{-5pt}
\includegraphics[width=0.99\linewidth]{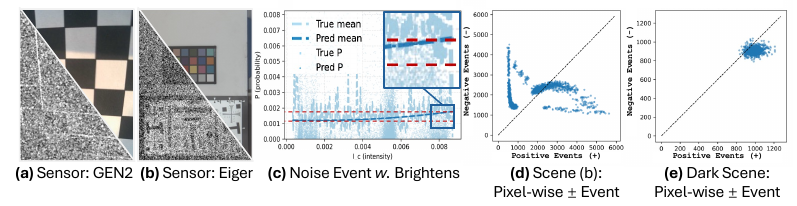}
\vspace{-10pt}
\caption{\small Noise events on GEN2 and Eiger. 
\textbf{(a,~b)} Event-probability  visualization from GEN2 and Eiger. 
\textbf{(c)} Event-probability \textit{vs.} brightness. 
\textbf{(d,~e)} Per-pixel positive \textit{vs.} negative event counts for an illuminated scene and for a dark scene.}
\vspace{-10pt}
\label{fig:6-EVS-Noise-Calibration-Shot}
\end{figure*}

\vspace{-3pt}
\section{Experiments\label{sec:downstream_tasks}}
\vspace{-3pt}
We conduct experiments in two stages: first, we validate the noise model and calibration framework on real data; second, we evaluate the simulator on real data—assessing temporal accuracy via video frame interpolation~\citeps{ma2025timelens,tulyakov2021time_1} and spatial fidelity via video deblurring~\cite{sun2022event,sun2023event}.
\textbf{Sensors:} We evaluate two hybrid event sensors\footnote{\url{ www.alpsentek.com/product}, \texttt{Eiger Model:APX003}}, GEN2 and Eiger, both using a Quad-Bayer~\citeps{gharbi2016deep} APS to increase spatial resolution.
Within each Quad-Bayer~\citeps{gharbi2016deep} block, GEN2 embeds one white event pixel, whereas Eiger integrates four color-filtered event pixels.
The APS resolution is $3246\times 2448$, and the EVS resolutions are $1632\times 1224$ and $816\times 612$, respectively.
\textbf{Dataset:}
To support both stages, we collect noise-calibration data, high-speed motion videos, and blurred videos, while additionally leveraging the publicly available datasets~\citeps{wu2025mipi,wu2024mipi,yunfan2024rgb}.
Each sensor outputs APS RAW frames and event streams.
\textbf{Implementation:}
All experiments are conducted with NumPy \citeps{harris2020array} and PyTorch \citeps{paszke2019pytorch} on a single NVIDIA A100 GPU.

\begin{figure*}[t!]
\centering
\includegraphics[width=0.96\linewidth]{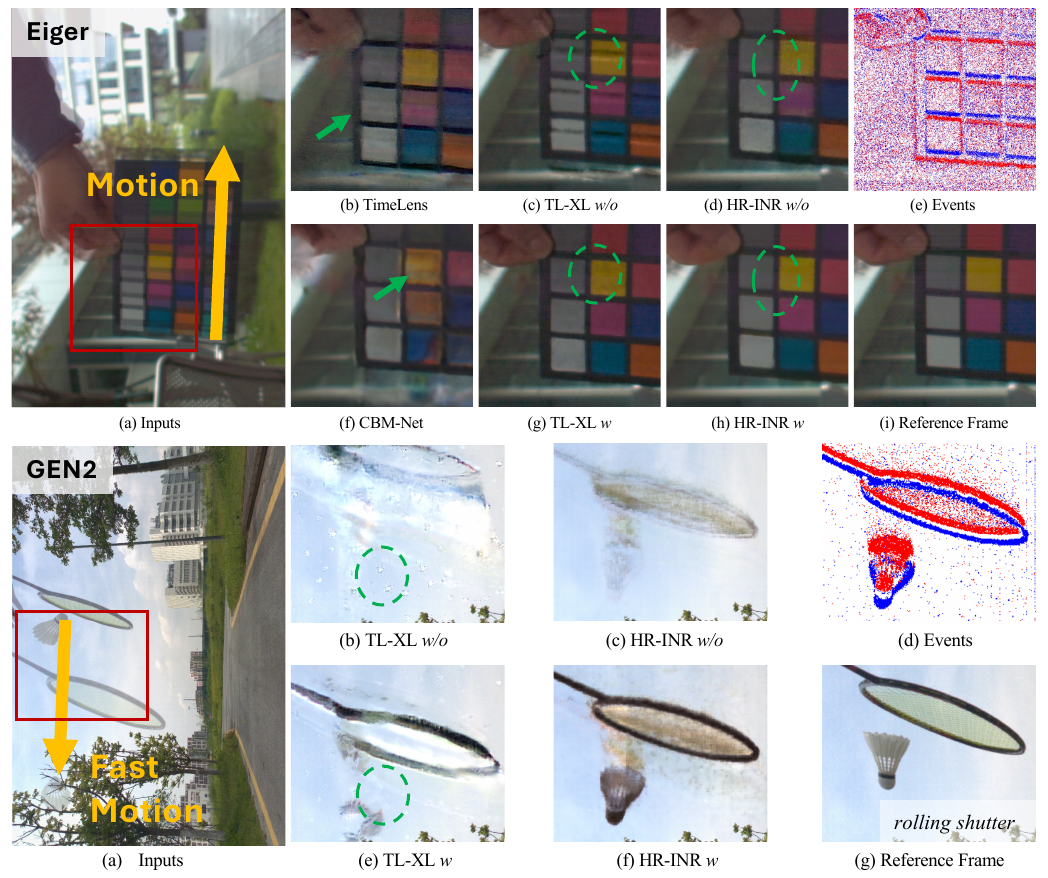}
\vspace{-7pt}
\caption{\small 
VFI under a skip-one-frame setting. The intermediate frame is removed and predicted from adjacent APS frames and events.
The reference frame uses rolling-shutter readout, which causes geometric distortion under fast motion.
\textit{w/o} and \textit{w} denote without and with H-ESIM fine-tuning, respectively.
}
\vspace{-22pt}
\label{fig:11-VFI-VIS}
\end{figure*}

\subsection{Noise Model and Calibration Analysis}
Following Sec.~\ref{sec:modeling} and Sec.~\ref{sec:calibration}, Fig.~\ref{fig:3-APS-Noise-Calibration-Light} and Fig.~\ref{fig:6-EVS-Noise-Calibration-Shot} present the noise-calibration results of APS and EVS.
\noindent\textbf{APS Noise Analysis:}
\textbf{(i)} \textit{Fixed-pattern noise:} Fig.~\ref{fig:3-APS-Noise-Calibration-Light} (a)–(b) show the fixed components.
The Eiger exhibits strong row noise, consistent with the influence of its embedded EVS pixels, one quarter per Quad-Bayer, on the readout path, whereas the GEN2 shows no visible banding.
Although fabrication and circuit design also contribute, the results indicate that stronger APS–EVS integration increases fixed-pattern noise.
\textbf{(ii)} \textit{Generated vs. real noise distributions:} Fig.~\ref{fig:3-APS-Noise-Calibration-Light} (c) shows that, for GEN2, the pixel-wise noise variance rises with image brightness. 
Fig.~\ref{fig:3-APS-Noise-Calibration-Light} (d) compare real and predicted noise from the clean intensity frame, showing a close statistical match.
\textbf{(iii)} \textit{Brightness-dependent noise:} Fig.~\ref{fig:3-APS-Noise-Calibration-Light} (e) plot real and predicted noise across brightness levels.
Furthermore, pixels under different color filters follow distinct trends, supporting the color-separated modeling.
In summary, two key insights emerge: \textbf{(1)} EVS–APS stacking and mixing amplify fixed noise; \textbf{(2)} color filters and pixel positions jointly influence per-pixel noise. These findings emphasize the complexity of hybrid-sensor noise modeling.

\begin{wrapfigure}{l}{0.37\linewidth}
\vspace{-7pt}
\centering
\includegraphics[width=\linewidth]{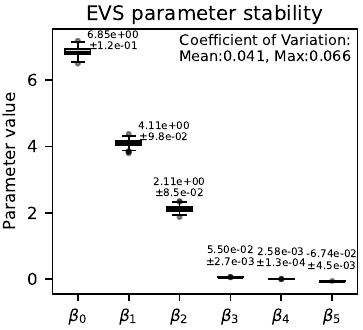}
\vspace{-18pt}
\caption{\small \label{fig:21-evs_beta_stability_nature_style}
Numerical stability of EVS parameter estimation.
We perform 200 independent runs with random initialization.
The boxplots summarize the distribution of the estimated parameters.
All coefficients of variation are below 0.066,
indicating strong stability across multi-start optimization.
}
\vspace{-40pt}
\label{fig:evs_stability}
\end{wrapfigure}
\noindent\textbf{EVS Noise Analysis:}
\textbf{(i)} \textit{Coupling of illumination and event noise:} Fig.~\ref{fig:6-EVS-Noise-Calibration-Shot}~(a)–(b) show that per-pixel event probability increases with brightness, confirming a direct link between illumination and noise events in Eq.~\ref{eq:positive_negative_possibilities}.
\textbf{(ii)} \textit{Event noise count distribution:} 
Fig.~\ref{fig:6-EVS-Noise-Calibration-Shot} (c) further shows that, as brightness increases, the entire distribution shifts upward, aligning with the prediction of Eq.~\ref{eq:event_estimation_final_noise}.
\textbf{(iii)} \textit{Coupling of polarity asymmetry and threshold}: 
Fig.~\ref{fig:6-EVS-Noise-Calibration-Shot}~(d)–(e) show that, in darkness, ON events slightly exceed OFF events, while under illumination, the ON/OFF balance varies with brightness, indicating nonlinear threshold–brightness coupling.
These observations confirm the illumination-dependent EVS noise captured by our model and calibration, and further reflect the intrinsic complexity of EVS noise behavior.
\textbf{(iv)} \textit{EVS Parameter Stability:} Fig.~\ref{fig:21-evs_beta_stability_nature_style} shows the convergence results under different random initializations. 
The maximum coefficient of variation across parameters is 0.066.
This indicates low dispersion relative to the mean and suggests stable convergence of the noise parameters.
\textit{See Appendix~\red{B} for more details.}

\begin{table}[t!]
\centering
\vspace{-9pt}
\setlength{\tabcolsep}{8.6pt}
\renewcommand{\arraystretch}{1.0}
\caption{\small
VFI results under the one-frame-skipping protocol on real data.
"Simulator" indicates the simulator used to generate the pre-training or finetune data for each model.
Methods marked \textit{w/o} and \textit{w} denote without and with H-ESIM fine-tuning.
}
\vspace{-9pt}
\resizebox{\linewidth}{!}{
\label{tab:vfi_psnr_ssim}
\begin{tabular}{c|r|rr|rr|rr}
\toprule
&Methods      & TL    & CBMNet   & TL-XL$~{\textit{w/o}}$ & TL-XL$~{\textit{w}}$ & HR-INR$~{\textit{w/o}}$ & HR-INR$~{\textit{w}}$ \\
\hline
&Simulator   & ESIM~\citeps{rebecq2018esim}  & ESIM~\citeps{rebecq2018esim}  & v2e~\citeps{hu2021v2e}                  & ours               & ESIM~\citeps{rebecq2018esim}                  & ours      \\
\midrule
\multirow{2}{*}{\rotatebox[origin=c]{90}{\textbf{Eiger}}}
& PSNR$\uparrow$ & 29.9948 & 31.9309 & 32.3046 & \gd{33.8743} & 32.4129 & \gd{35.2425} \\
& SSIM$\uparrow$ & 0.5793 & 0.8410 & 0.7755 & \gd{0.8629} & 0.7821 & \gd{0.8862} \\
& LPIPS$\downarrow$ & 0.4042 & 0.3021 & 0.2143 & \gd{0.1677} & 0.1979 & \gd{0.0787} \\
\midrule
\multirow{2}{*}{\rotatebox[origin=c]{90}{\textbf{GEN2}}} 
&PSNR$\uparrow$ & 31.6309 & 33.7758 & 34.1787 & \gd{34.5801} & 34.2631 & \gd{35.5198}   \\
&SSIM$\uparrow$ &  0.7583 &  0.8329 &  0.9112 &  \gd{0.9127} &  0.9134 &  \gd{0.9278} \\
& LPIPS$\downarrow$ & 0.2175 & 0.2643 & 0.0865 & \gd{0.0535} &  0.0726 &  \gd{0.0419} \\
\bottomrule
\end{tabular}
}
\vspace{-10pt}
\end{table}

\begin{table}[t]
\centering
\setlength{\tabcolsep}{13.6pt}
\renewcommand{\arraystretch}{1.0}
\caption{\small Spatial deblurring on the hybrid sensor Eiger with RAW and events.
Methods marked \textit{(w/o)} are trained only on the dataset~\citeps{wu2025mipi}, while those marked \textit{(w)} are additionally fine-tuned on \textbf{H-ESIM}.
\gd{blue} is improvement and \bd{red} is degradation.}
\vspace{-8pt}
\label{tab:spatial_interp_metrics}
\resizebox{\linewidth}{!}{
\begin{tabular}{l|l|rrr}
\toprule
\textbf{Method} &
\textbf{Simulator} &
\textbf{CLIP-IQA}~$\uparrow$ &
\textbf{MUSIQ}~$\uparrow$ &
\textbf{NRQM}~$\uparrow$ \\
\midrule
eSL~\cite{yu2023learning}~\textit{(w/o)} & \citeps{wu2025mipi}
& 0.3070 & 18.74 &   4.091\\
\rowcolor{gray!10}
eSL~\cite{yu2023learning}~\textit{(w)} & \citeps{wu2025mipi} + H-ESIM
& \gd{0.4346} & \gd{22.97}   & \gd{5.899}\\
MAER~\cite{sun2024motion}~\textit{(w/o)} & \citeps{wu2025mipi}
& 0.3297 & 18.88   & 5.038\\
\rowcolor{gray!10}
MAER~\cite{sun2024motion}~\textit{(w)} & \citeps{wu2025mipi} + H-ESIM
& \gd{0.3370} & \gd{19.05}   & \gd{5.064}\\
EFNet~\cite{sun2022event}~\textit{(w/o)}& \citeps{wu2025mipi}
& 0.2208 & 16.83   & 3.797\\
\rowcolor{gray!10} 
EFNet~\cite{sun2022event}~\textit{(w)} & \citeps{wu2025mipi} + H-ESIM
& \gd{0.4248} & \gd{21.52}   & \gd{5.108} \\
\bottomrule
\end{tabular}
}
\vspace{-10pt}
\end{table}

\subsection{Temporal and Spatial Evaluation of H-ESIM}
We evaluate temporal and spatial modeling on dynamic scenes captured by GEN2 and Eiger.
\noindent\textbf{Temporal Analysis:}
For temporal evaluation, we benchmark TimeLens~\citeps{tulyakov2021time}, CBMNet~\citeps{kim2023event}, TimeLens-XL~\citeps{ma2025timelens}, and HR-INR~\citeps{lu2024hr} on VFI. 
Official training code is available for TL-XL and HR-INR, which we fine-tune using data generated by H-ESIM.
In testing, we remove one intermediate APS frame and predict it from two adjacent frames and events. 
We report reference-based metrics, including PSNR, SSIM~\citeps{wang2004image}, and LPIPS~\citeps{zhang2018unreasonable}.
The reference APS frame is captured with rolling-shutter readout and exhibit geometric distortion under fast motion; it therefore serves as a reference rather than a distortion-free ground truth.
\noindent{\textit{(i) Quantitative Results:}}
Tab.~\ref{tab:vfi_psnr_ssim} reports PSNR, SSIM, and LPIPS.
After fine-tuning with H-ESIM, both TimeLens-XL and HR-INR show consistent improvements across distortion-based and perceptual metrics.
The gains are more evident on Eiger, where fixed-pattern noise and layout-induced artifacts are stronger.
This suggests that the calibrated simulator provides training data that better matches real hybrid sensor statistics, thereby reducing the domain gap between synthetic and real data.
\noindent{\textit{(ii) Qualitative Results:}}
Fig.~\ref{fig:11-VFI-VIS} presents representative visual comparisons.
Baseline models exhibit edge misalignment in moving regions.
TL-XL without fine-tuning occasionally amplifies event fixed noise, leading to isolated noise points in reconstructed areas.
After H-ESIM fine-tuning, such artifacts are suppressed, and reconstructed edges appear sharper and more coherent.
Under extremely fast motion, none of the methods fully recover the exact object contours due to rolling-shutter distortion and event sparsity. 
However, fine-tuned models produce clearer edge transitions and better structural continuity.
In particular, HR-INR more reliably reconstructs the contour of the fast-moving shuttlecock, reflecting improved temporal consistency under hybrid-sensor noise.

\noindent\textbf{Spatial Analysis:}
We further assess spatial modeling through hybrid-sensor deblurring.
Since ground-truth sharp frames are unavailable, we report no-reference image quality metrics:
CLIP-IQA~\citeps{wang2023exploring} and MUSIQ~\citeps{ke2021musiq} for perceptual and semantic fidelity,
BRISQUE~\citeps{mittal2012no} for naturalness and distortion, and
We evaluate eSL~\citeps{yu2023learning}, MAER~\citeps{sun2024motion}, and EFNet~\cite{sun2022event} under the same setting as~\citeps{wu2025mipi}, comparing models fine-tuned with and without H-ESIM.
Tab.~\ref{tab:spatial_interp_metrics} shows that fine-tuning on H-ESIM leads to consistent improvements in perceptual metrics across all models. 
Gains are most evident in CLIP-IQA, with similar trends observed in MUSIQ and NRQM. 
These results suggest that calibrated simulation helps reduce spatial artifacts and improves structural coherence under real hybrid-sensor noise.
Overall, H-ESIM fine-tuning improves spatial detail recovery and perceptual consistency on real hybrid-sensor data. \textit{See Appx.~\red{D} for more visualization.}

\noindent\textbf{Comparison with Generic Simulators:}
Tab.~\ref{tab:vfi_psnr_ssim} and Tab.~\ref{tab:spatial_interp_metrics} list the simulator used during training. Generic simulators such as ESIM~\citeps{rebecq2018esim} and v2e~\citeps{hu2021v2e} generate contrast-based events but do not model hybrid-sensor-specific noise, including layout-dependent fixed noise.
After fine-tuning with H-ESIM, both VFI and deblurring models show consistent improvements. This indicates that calibrating simulator statistics to real hybrid sensors reduces the synthetic–real domain gap and improves robustness to hybrid-specific noise.

\vspace{-4pt}
\section{Conclusion}
\vspace{-3pt}
This work presents the first simulation framework for hybrid event sensors.
We develop a unified imaging noise model, a calibration method, and a realistic simulator, H-ESIM, supported by comprehensive temporal and spatial analyses for downstream tasks.
Our framework establishes a foundation for future hybrid-sensor research and aims to inspire broader community efforts in hybrid event sensor.
\noindent\textbf{Limitation:}
As an initial step toward hybrid-sensor simulation, our model does not explicitly cover extreme conditions, such as very low illumination, extreme temperatures, or bandwidth constraints. Under these regimes, noise statistics may deviate from the Gaussian approximation adopted in this work. Extending the model to account for such conditions remains future work.
\\

{\small \noindent\textbf{Acknowledgements:} This work was supported in part by the National Natural Science Foundation of China (Grant No. 92370204), in part by the National Key R\&D Program of China (Grant No. 2023YFF0725001), in part by Guangdong Provincial Key Laboratory of Frontier Basic Science for All-domain Intelligence, in part by the Guangdong Basic and Applied Basic Research Foundation (Grant No. 2023B1515120057), in part by the Key-Area Special Project of Guangdong Provincial Ordinary Universities (2024ZDZX1007).}

{
    \small
    \bibliographystyle{splncs04}
    \bibliography{ref}
}
\clearpage

\end{document}